\documentclass[letterpaper]{article} 
\usepackage[draft]{aaai2026}  
\usepackage{times}  
\usepackage{helvet}  
\usepackage{courier}  
\usepackage[hyphens]{url}  
\usepackage{graphicx} 
\usepackage{natbib}  
\usepackage{caption} 
\usepackage{algorithm}
\usepackage{algorithmic}

\usepackage{amsmath}
\usepackage{enumitem}
\usepackage{amsfonts}
\usepackage{subcaption} 
\usepackage{booktabs}
\usepackage{multirow}

\usepackage{newfloat}
\usepackage{listings}
\DeclareCaptionStyle{ruled}{labelfont=normalfont,labelsep=colon,strut=off} 
\floatstyle{ruled}
\newfloat{listing}{tb}{lst}{}
\floatname{listing}{Listing}
\title{Are Today’s LLMs Ready to Explain Well-Being Concepts?}
\author{
    Bohan Jiang\textsuperscript{\rm 1},
    Dawei Li\textsuperscript{\rm 1},
    Zhen Tan\textsuperscript{\rm 1},
    Chengshuai Zhao\textsuperscript{\rm 1},
    Huan Liu\textsuperscript{\rm 1}
}
\affiliations{
    \textsuperscript{\rm 1}School of Computing and Augmented Intelligence, Arizona State University, USA\\

    \{bjiang14, daweili5, ztan36, czhao93, huanliu\}@asu.edu
}

\usepackage{bibentry}

\begin{document}

\maketitle

\begin{abstract}

Well-being encompasses mental, physical, and social dimensions essential to personal growth and informed life decisions. As individuals increasingly consult Large Language Models (LLMs) to understand well-being, a key challenge emerges: Can LLMs generate explanations that are not only accurate but also tailored to diverse audiences? High-quality explanations require both factual correctness and the ability to meet the expectations of users with varying expertise. In this work, we construct a large-scale dataset comprising 43,880 explanations of 2,194 well-being concepts, generated by ten diverse LLMs. We introduce a principle-guided LLM-as-a-judge evaluation framework, employing dual judges to assess explanation quality. Furthermore, we show that fine-tuning an open-source LLM using Supervised Fine-Tuning (SFT) and Direct Preference Optimization (DPO) can significantly enhance the quality of generated explanations. Our results reveal: (1) The proposed LLM judges align well with human evaluations; (2) explanation quality varies significantly across models, audiences, and categories; and (3) DPO- and SFT-finetuned models outperform their larger counterparts, demonstrating the effectiveness of preference-based learning for specialized explanation tasks.

\end{abstract}

\section{Introduction}\label{sec:intro}
Well-being is a multi-dimensional concept without a single clear and universally accepted definition~\cite{alexandrova2017philosophy}. In general, people describe well-being as ``how people feel and how they function both on a personal and social level, and how they evaluate their lives as a whole,'' pointing to a complex interplay of mental, physical, and social dimensions~\cite{topp20155}. Gaining a clear understanding of well-being concepts is vital for self-reflection, decision-making, and personal growth~\cite{diener2000subjective}. 

Recent Large Language Models (LLMs) are increasingly becoming primary sources of knowledge for individuals seeking insights on well-being and its related concepts~\cite{xiong2024search, wu2024like}. As users turn to LLMs for such guidance, the quality of the explanations they receive plays a critical role. However, generating high-quality explanations for a well-being concept is a challenging task. A good explanation requires more than just factual accuracy; it must be tailored to the user's specific needs and level of expertise~\cite{cho2018writing}. Due to the knowledge gap between domain experts and the general public, it is difficult to find a one-size-fits-all explanation~\cite{keil2006explanation}. For example, a layperson requires accessible language, real-world examples, and actionable advice. In contrast, a domain expert would prefer technical terminology, critical nuance, and evidence-based substantiation~\cite{jarden2023wellbeing}. The unexamined quality of LLM-generated explanations, coupled with the difficulty of the task, presents a significant research challenge. In this paper, we pioneer the exploration of the following \textbf{\textit{Research Question:}} Are Today’s LLMs Ready to Explain Complex Well-Being Concepts?

\begin{figure}[t]
  \centering
  \includegraphics[width=.99\columnwidth]{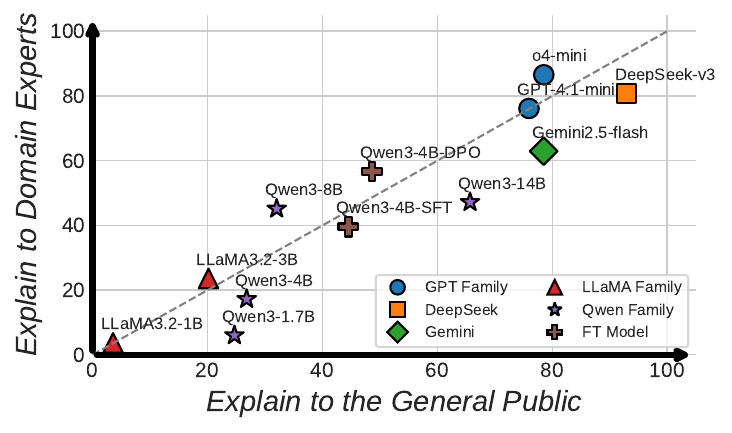}
  \caption{Illustration of the explanation capability of LLMs.}
  \label{fig:2d}
\end{figure}

\begin{figure*}[ht]
  \centering
  \includegraphics[width=.99\linewidth]{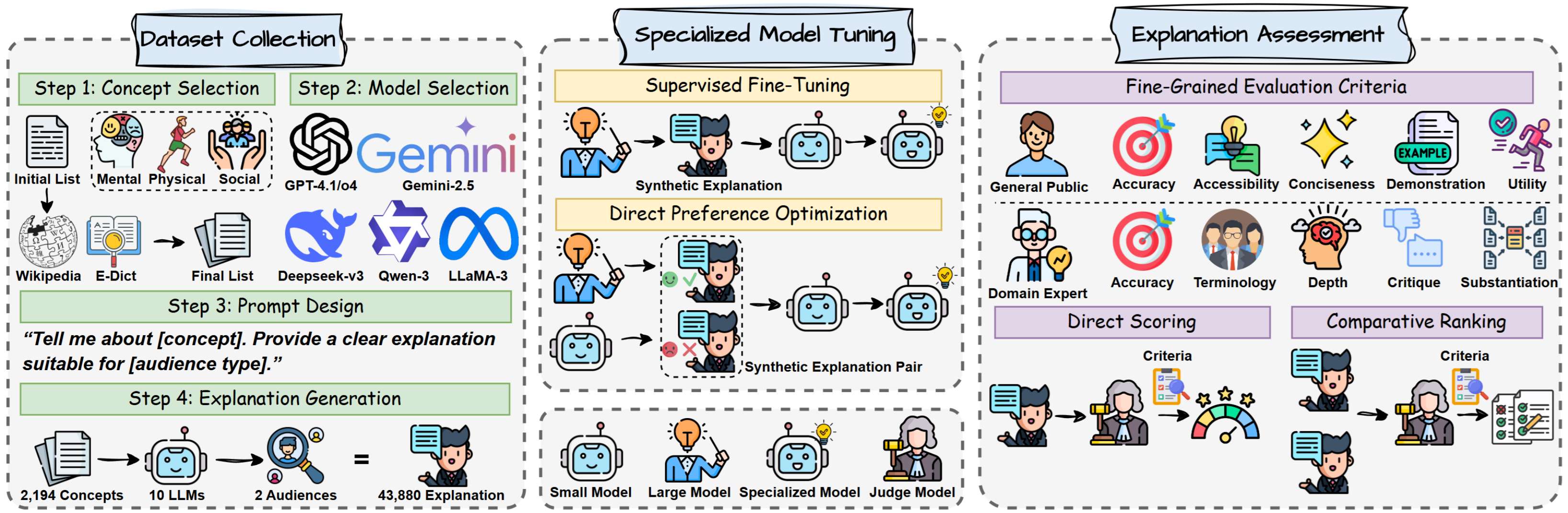}
  \caption{Overview of the research pipeline.}
  \label{fig:pipeline}
\end{figure*}

To address this, our work provides the first large-scale, systematic investigation of existing LLMs' capabilities in explaining well-being concepts. We follow a comprehensive research pipeline (Figure~\ref{fig:pipeline}), beginning with the curation of a \textbf{large-scale dataset}, collecting 43,880 explanations from 10 diverse LLMs for 2,194 concepts. Those concepts are chosen from well-being-related literature~\cite{diener2000subjective, topp20155, tov2018well}. We then propose a novel \textbf{evaluation framework} that adapts the principle-guided LLM-as-a-judge paradigm~\cite{zheng2023judging}, using two distinct judge models guided with fine-grained, audience-level principles to assess explanation quality. Finally, we investigate \textbf{pathways for improvement} by fine-tuning an open-source model using both Supervised Fine-Tuning (SFT) and Direct Preference Optimization (DPO)~\cite{rafailov2023direct} to create specialized, high-performing explanation models.

Our empirical results first validate that the principle-guided evaluation framework provides reliable judgments that align with human evaluators. Our analysis of the 10 baseline LLMs reveals significant performance disparities, with larger models consistently outperforming smaller ones (Figure~\ref{fig:2d}), but even top models, such as o4-mini~\cite{jaech2024openai} and Gemini-2.5-flash~\cite{comanici2025gemini}, exhibit shared weaknesses in providing practical advice and nuanced analysis. We also find that generating explanations for domain experts is particularly challenging (Figure~\ref{fig:2d}), leading to a higher likelihood of factual inaccuracies. Crucially, we demonstrate that both SFT and DPO substantially improve the capabilities of a smaller model, with the DPO-tuned model's performance surpassing that of its much larger variants, proving the value of our curated preference data.
In summary, our main contributions are as follows:
\begin{itemize}[leftmargin=*,itemsep=1pt]
  \item \textbf{Novel Datasets:} We develop the first well-being concept explanation dataset. It consists of 43,880 LLM-generated concept explanations for 2,194 distinct mental, physical, and social well‐being concepts. We also provide audience-aware, specific fine-tuning datasets for SFT and DPO.
  
  \item \textbf{Fine-Grained Evaluation:} We propose a LLM‐as-a-judge framework with fine-grained audience-level principles as guidance. We evaluate each concept explanation using both direct scoring and comparative ranking. We also adopt a co-judge strategy to mitigate evaluation bias. 
  
  \item \textbf{Empirical Experiments:} We conduct comprehensive experiments on ten pre-trained and two fine-tuned LLMs. We reveal the nuanced performance differences among these models. We demonstrate the explanation quality improvements of fine-tuned models over larger baselines.
  
  \item \textbf{Practical Implications:} We provide in-depth analyses to probe model size effects, audience effects, and principle‐wise variation. We rank LLMs per audience-level principle and point out the common weaknesses of LLMs. 
\end{itemize}

\section{Related Work}
\subsection{LLMs for Well-Being}
LLMs are increasingly being developed as proactive agents to promote human well-being~\cite{lin2020caire, chen2024structured, reategui2025llm}. A major line of research involves intelligent chatbots designed to address the shortage of conventional mental health services by offering scalable and effective solutions, from initial diagnoses to follow-up support in clinical domains~\cite{prakash2020intelligent, jo2023understanding, nie2024llm}. In the educational area, similar chatbot technologies are used to enhance student well-being by serving as intelligent teaching assistants that improve the learning experience, answer queries, and support student success~\cite{chae2023tutoring, grossman2019mathbot, gao2025agent4edu}.
Beyond direct user support, another body of work utilizes LLMs as protective safeguards for societal well-being. This research focuses on combating the negative psychological and social impacts of harmful online content. LLMs are being deployed to detect and mitigate misinformation~\cite {chen2023can, hu2024bad}, disinformation~\cite{jiang2024disinformation, zhang2025llms}, and hate speech~\cite{shen2025hatebench, meguellati2025llm}, thereby aiming to create safer digital environments.
While previous work focuses on using LLMs to promote or protect well-being, it presupposes that these models have a coherent grasp of the concept itself. There is a lack of research investigating whether LLMs can correctly understand and articulate the nuances of well-being concepts. This work aims to address this gap by systematically evaluating LLM-generated well-being concept explanations.


\subsection{Evaluation of LLM-Generated Content}
\noindent\textbf{Traditional Assessment Metric:} Traditional metrics like BLEU~\cite{papineni2002bleu} and ROUGE~\cite{lin2004rouge} rely heavily on exact matching to evaluate models' generation quality. Subsequent methods, such as BERTScore~\cite{zhangbertscore} and BARTScore~\cite{yuan2021bartscore}, improve upon this by using contextual embeddings, but remain incapable of capturing nuanced features~\cite{post2018call}.

\noindent\textbf{LLM-as-a-judge:} The advanced capabilities of LLMs have inspired a paradigm shift towards dynamic reference-free assessment~\cite{wang2023chatgpt}. LLM-as-a-judge, as one of the leading evaluation paradigms, has been widely adopted due to its ability to conduct nuanced evaluations like humans~\cite{zheng2023judging,li2024llmasajudge}. It has been used in domains like academic writing~\cite{liu2023reviewergpt}, code generation~\cite{mcaleese2024llm}, and social science~\cite{jiang2024assessing}, to evaluate the quality of LLM-produced open-ended generation. However, recent studies have revealed various biases and vulnerabilities of the LLM-as-a-judge paradigm, raising concerns in this technique~\cite{li2025preference}.

\noindent\textbf{Principle-Guided Evaluation:} To address these limitations, researchers proposed principle-guided evaluation with LLM-as-a-judge~\cite{li2024llmasajudge}, where a set of comprehensive and well-designed rules or rubrics is given to the LLM judge for improving the assessment's fairness and reliability. Following studies further improve it by providing domain~\cite{ye2023flask} or sample-level principles~\cite{kim2025biggen,gunjal2025rubrics,viswanathan2025checklists}, instructing LLM judges with more fine-grained guidelines. Building on this line of work, we introduce \textit{audience-level principles}: tailored guidelines that align the judge’s perspective with the needs of distinct explanation audience groups.
(e.g., general public and domain experts).

\section{Methods}\label{sec:method}
\subsection{Collecting Concept Explanation Dataset}
To systematically evaluate the quality of LLM-generated explanations for well-being concepts, we conduct a rigorous data collection procedure comprising the following steps:

\paragraph{Step 1: Well-Being Concept Selection.}
We start by compiling a comprehensive list of well-being concepts across three primary dimensions: mental, physical, and social well-being. Initial concepts are identified based on their relevance, popularity, and coverage in related literature on human well-being~\cite{diener2000subjective, topp20155, tov2018well}. We further expand this list through cross-referencing synonyms and related terms from Wikipedia and the Oxford English Dictionary. The final dataset consists of 451 mental, 1,011 physical, and 732 social well-being concepts.

\paragraph{Step 2: Generation Model Selection.}
We select ten diverse large language models (LLMs) for generating concept explanations. These include four larger API-based proprietary models known for their advanced capabilities:
\begin{itemize}[leftmargin=*,itemsep=1pt]
    \item GPT-4.1-mini~\cite{achiam2023gpt}.
    \item OpenAI-o4-mini~\cite{jaech2024openai}.
    \item Gemini-2.5-flash~\cite{comanici2025gemini}.
    \item Deepseek-V3~\cite{liu2024deepseek}.
\end{itemize}
Additionally, six smaller open-source LLMs are included:
\begin{itemize}[leftmargin=*,itemsep=1pt]
    \item Qwen-3 (1.7B, 4B, 8B, and 14B)~\cite{yang2025qwen3}.
    \item LLaMA-3.2-instruct (1B and 3B)~\cite{grattafiori2024llama}.
\end{itemize}
This combination provides comprehensive coverage across different scales, architectures, and training paradigms.
\paragraph{Step 3: Generation Prompt Design.}
To ensure consistency and emulate realistic user-LLM interactions, we design a standardized prompt template:
\begin{quote}
    \textit{``Tell me about [concept]. Provide a clear explanation suitable for [audience type].''}
\end{quote}
We iteratively refine this template through pilot testing, using two audience categories: ``general public'' and ``domain experts', to guide LLMs in generating targeted explanations.

\paragraph{Step 4: Concept Explanation Generation.}
Applying the finalized prompt template, we query each of the 2,194 concepts against all 10 selected LLMs, resulting in a total of 43,880 concept explanations (2,194 concepts × 10 LLMs × 2 audience types). To minimize variability and randomness in model outputs, all generations are conducted using a deterministic setting with LLMs' $\text{temperature} = 0$.

\subsection{Fine-Tuning Specialized Model}
To validate whether the collected dataset is suitable for fine-tuning a specialized model for better well-being concept explanation, we investigate two distinct fine-tuning strategies: \textbf{Supervised Fine-Tuning (SFT)} and \textbf{Direct Preference Optimization (DPO)}. These methods are applied separately to the same pre-trained base model (Qwen-3-4B). In both strategies, we denote the model being trained as $M_{\theta}$.

\subsubsection{Supervised Fine-Tuning.}
SFT aims to adapt the base model $M_{\theta}$ to generate outputs that conform to the format and style of high-quality explanations.

\paragraph{Step 1: SFT Data Preparation.} We construct our SFT dataset $D_{SFT}$ by applying a filtering process to each well-being concept explanation to select high-quality responses. We first make an assumption based on previous work -- for a given prompt $P$, the quality of the response from larger LLMs generally outperforms those from the smaller LLMs~\cite{askell2021general, kim2023aligning}. Therefore, an explanation $E_{i,j,k}$ (for concept $c_i$, generation model $M_j$, and audience $a_k$) is included in $D_{SFT}$ only if it is generated by a larger LLM (e.g., Gemini-2.5-flash). 

\paragraph{Step 2: SFT Objective.} The base model $M_{\theta}$ is then fine-tuned on the curated dataset $D_{SFT}$. For each concept, we use the standardized prompt template containing the concept $c_i$ and audience type $a_k$ as the input prompt $P_i$ and the corresponding high-quality explanation as the target output $E_i$. The SFT objective is to train $M_{\theta}$ by minimizing the negative log-likelihood loss $\mathcal{L}_{SFT}(\theta)$:
\begin{equation}
\mathcal{L}_{SFT}(\theta) = - \sum_{\substack{(P_i, E_i) \\ \in D_{SFT}}} \sum_{t=1}^{|E_i|} \log P(E_{i,t} | P_i, E_{i,<t}; \theta)
\end{equation}
where $E_{i,t}$ is the $t$-th token of the target explanation $E_i$.

\subsubsection{Direct Preference Optimization.}
DPO directly optimizes the model based on human preference data~\cite{rafailov2023direct}. It is designed to explicitly teach the model to distinguish between high-quality and low-quality responses.

\paragraph{Step 1: DPO Data Preparation.} The preference dataset $D_{DPO}$ consists of pairs of preferred and dispreferred responses for each input prompt $P$. Similar to the way we construct $D_{SFT}$, we create a pool of \textit{good} and \textit{bad} explanations for every well-being concept in our collected data:
\begin{itemize}[leftmargin=*,itemsep=1pt]
    \item \text{Good Explanation ($E_g$)}: it only includes well-being concept explanations generated by the larger LLMs.
    \item \text{Bad Explanation ($E_b$)}: it only includes well-being concept explanations generated by the smaller LLMs.
\end{itemize}
As a result, the final dataset $D_{DPO}$ is composed of multiple $(P, E_g, E_b)$ for each concept and audience type.

\paragraph{Step 2: DPO Objective.} The policy model $M_{\theta}$ is optimized by DPO to increase the likelihood of explanation $E_g$ over $E_b$. This is guided by a frozen reference model $M_{ref}$, which is the original pre-trained base model $M_{\theta}$. The DPO loss function is defined as:
\begin{equation}
\mathcal{L}_{DPO}(\theta) = -\mathbb{E}_{\substack{(P,E_g,E_b) \\ \in D_{DPO}}\!}\left[\log\sigma\!\left(\!\beta\!\left(\!\log\tfrac{\pi_\theta^g}{\pi_{ref}^g} - \log\tfrac{\pi_\theta^b}{\pi_{ref}^b}\right)\!\!\right)\right]
\end{equation}
where $\pi_\theta^g \equiv \pi_\theta(E_g|P)$, $\pi_{ref}^b \equiv \pi_{ref}(E_b|P)$, $\sigma$ is the sigmoid function, and $\beta$ controls deviation from $\pi_{ref}$.

\subsection{Assessing Concept Explanation Quality}
We employ a principle-guided \textbf{LLM-as-a-judge} paradigm to assess explanation quality, leveraging two powerful Large Reasoning Models (LRMs) as judges, $J = \{J_1, J_2\}$, where $J_1$ is Gemini-2.5-Pro~\cite{comanici2025gemini} and $J_2$ is DeepSeek-R1~\cite{guo2025deepseek}. Judges will assess the quality using \textit{Direct Scoring} and \textit{Comparative Ranking} based on the predefined audience-level principles. 

\paragraph{Step 1: Fine-Grained Evaluation Criteria.}
To enhance consistency and interpretability of the evaluation process, inspired by previous work~\cite{ye2023flask}, we carefully define evaluation criteria with fine-grained, \textit{audience-level principles} tailored to two types of audiences. For the \textit{general public} without sufficient domain knowledge:
\begin{itemize}[leftmargin=*,itemsep=1pt]
    \item Accuracy: Provide an accurate definition of the concept.
    \item Accessibility: Use of simple, everyday language.
    \item Conciseness: Brief and direct explanations without unnecessary verbosity.
    \item Demonstration: Use of relatable analogies, stories, or real-world examples.
    \item Utility: Provision of actionable and practical advice.
\end{itemize}
For \textit{domain experts} with sufficient domain knowledge:
\begin{itemize}[leftmargin=*,itemsep=1pt]
    \item Accuracy: Provide an accurate definition of the concept.
    \item Terminology: Use of professional, field-specific jargon.
    \item Depth: Comprehensive and nuanced analysis of concepts.
    \item Critique: Identification of limitations and controversies.
    \item Substantiation: Inclusion of references and citations from peer-reviewed literature.
\end{itemize}
Note that \textit{Accuracy} is presented in both scenarios because of its importance for analyzing any inaccurate or hallucinated definition in the generated concept explanation. 

\paragraph{Step 2: Direct Scoring.}
In this method, judges assign a score to each explanation \textit{per principle}. For a given concept explanation $E_{i,j,k}$ (for concept $c_i$, generation model $M_j$, and audience $a_k$), each judge $J_l$ provides a score $S_l(E_{i,j,k}, v) \in [1, 5]$ for each principle $v$.
The final score for an explanation on a specific principle $v$ is the average score from $J_1$ and $J_2$:
\begin{equation}
S(E_{i,j,k}, v) = \frac{1}{|J|} \sum_{l=1}^{|J|} S_l(E_{i,j,k}, v).
\end{equation}
To assess a model's performance on a specific principle for a given audience, we aggregate the scores $S(E_{i,j,k}, v)$ across all concepts. The total quality score for model $M_j$ on principle $v$ for audience $a_k$ is:
\begin{equation}
Q_{DS}(M_j, k, v) = \sum_{i=1}^{|C|} S(E_{i,j,k}, v).
\end{equation}
Note that LLMs assign \textit{Accuracy} scores for the generated explanation by comparing the definition from Wikipedia and Dictionary as ground truth (i.e., the maximum score).

\paragraph{Step 3: Comparative Ranking.}
In this method, judges compare each generated explanation against a baseline reference $E_{ref}$ (i.e., Qwen-3-14B) \textit{per principle}. For each principle $v$, the comparison yields an outcome $O(E_{i,j,k}, v) \in \{\text{win, loss, tie}\}$.
A conflict between judges on any given principle (e.g., $J_1$ outputs \textit{win} and $J_2$ outputs \textit{loss}) will result in a \textit{tie} for that specific principle.

A model's performance is then quantified by its \textbf{win rate} for each principle and each audience type. The win rate for model $M_j$ on principle $v$ for audience $a_k$ is calculated as:
\begin{equation}
W(M_j, k, v) = \frac{|\{E_{i,j,k} \mid O(E_{i,j,k}, v) = \text{win}\}|}{|C|}
\end{equation}
For \textit{Accuracy}, the judge assigns an outcome based on which explanation ($E_{ref}$ and $E_{i,j,k}$) is closer to the ground-truth definition. For example, if the baseline reference's explanation $E_{ref}$ is closer to Wikipedia's definition, the final outcome will be \textit{loss}.


\begin{figure*}[ht]
  \centering
  \includegraphics[width=.96\linewidth]{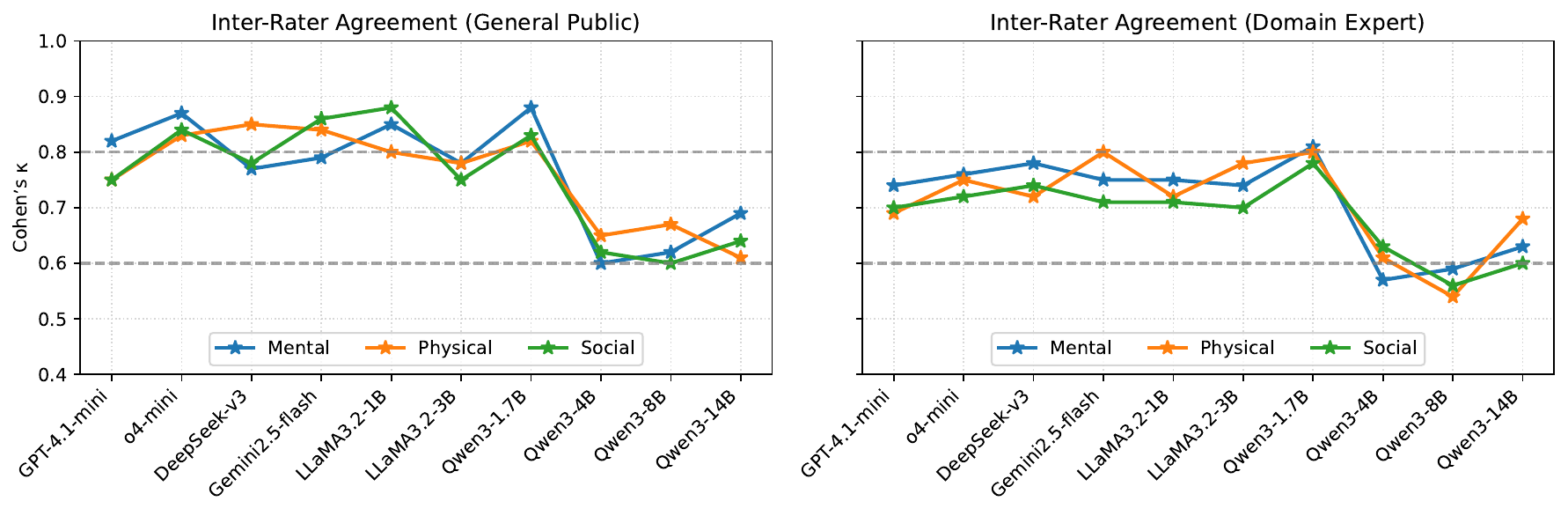}
  \caption{Cohen’s kappa scores between LLM‐as‐a-judge and human annotators. Dashed lines at 0.6 and 0.8 indicate substantial (0.61 to 0.80) and almost‐perfect (0.81 to 1) agreement, respectively.}
  \label{fig:kappa_line}
\end{figure*}

\section{Results}
In this section, we present our empirical results to answer the following research questions:
\begin{itemize}[leftmargin=*,itemsep=1pt]
    \item \textbf{RQ1}: Does the proposed principle-guided LLM-as-a-judge framework provide human-level evaluation?
    \item \textbf{RQ2}: How do the capabilities of LLMs differ when explaining well-being concepts in different scenarios?
    \item \textbf{RQ3}: To what extent can fine-tuning via SFT and DPO improve LLMs' well-being concept explanation abilities?
\end{itemize}

\subsection{Validations of our evaluation framework (RQ1)}
To validate the reliability of our principle-guided LLM-as-judge paradigm and answer \textbf{RQ1}, we conduct human evaluations to assess the explanations using the comparative ranking strategy with identical evaluation principles. Specifically, we compare the LLMs' judgements against human annotations on a held-out set of 50 explanations per model. We compute the overall win rate for each LLM by calculating the average win rate among all audience-level principles. The inter-rater agreement is measured using Cohen’s kappa~\cite{cohen1960coefficient} on the mental, physical, and social well-being concept categories. We report the results for general public and domain expert explanations separately. Figure~\ref{fig:kappa_line} visualizes the level of agreement for all evaluated LLMs. We observe that LLM-as-a-judge is more reliable when evaluating concept explanations for the general public, evidenced by overall higher Cohen's kappa scores. Moreover, they have \textbf{more agreement on evaluation results from larger LLMs and extremely smaller LLMs} (LLaMA-3.2-1B, 3B, and Qwen-3-1.7B). This indicates that it is easy for LLMs to recognize extremely \textit{good} and \textit{bad} well-being concept explanations. However, their judgments become slightly unreliable (i.e., moderate to substantial agreement) when dealing with relatively moderate quality explanations. Another finding is that there is no significant inter-rater agreement discrepancy between the three well-being categories.

\subsection{Differences in Well-being Concept Explanation Capability (RQ2)}

\begin{figure*}[ht]
  \centering
  \begin{subfigure}[b]{0.5\textwidth}
    \centering
    \includegraphics[width=\textwidth]{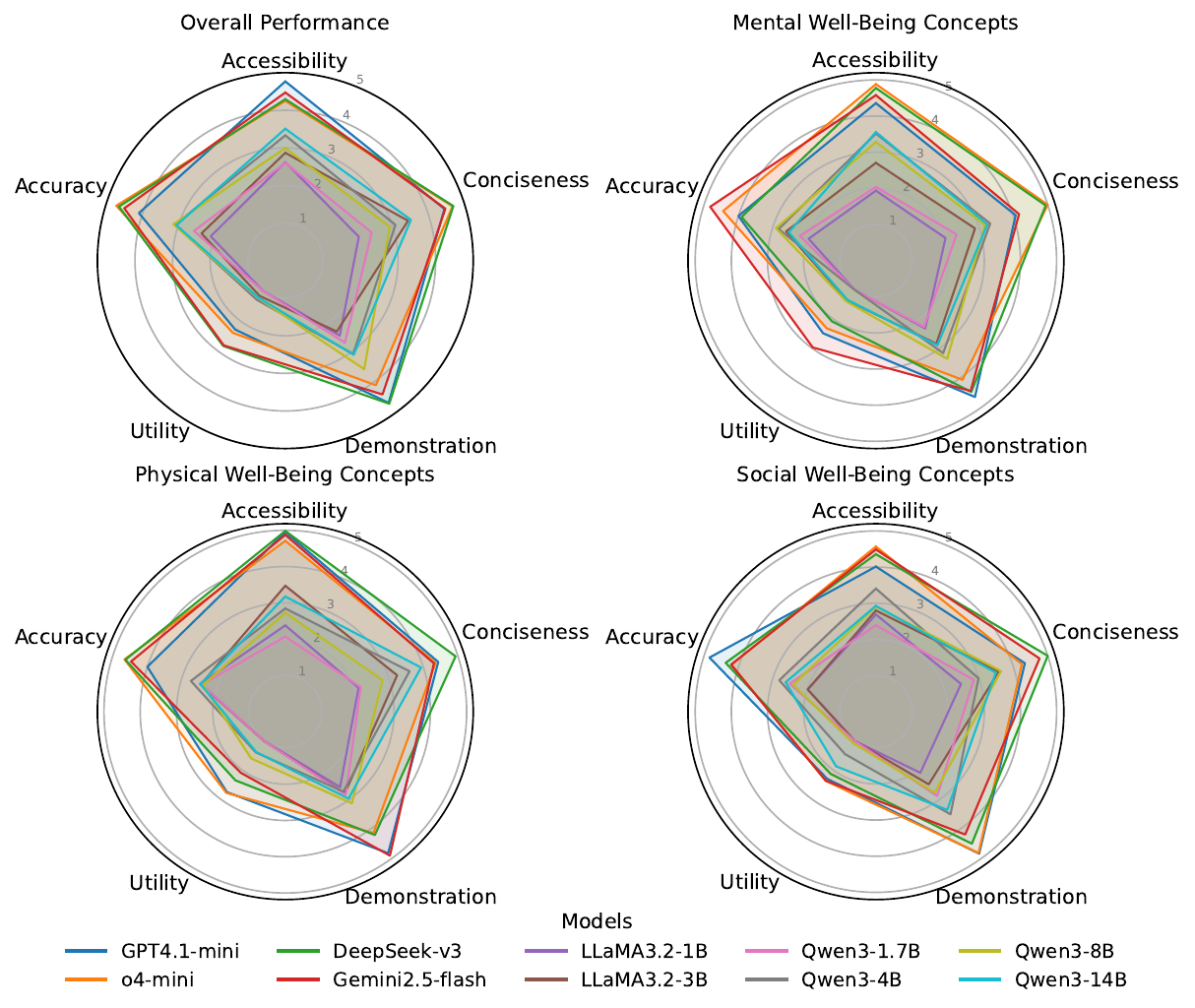}
    \caption{Scores of concept explanations for the general public.}
    \label{fig:radar_ne}
  \end{subfigure}
  \hfill
  \begin{subfigure}[b]{0.47\textwidth}
    \centering
    \includegraphics[width=\textwidth]{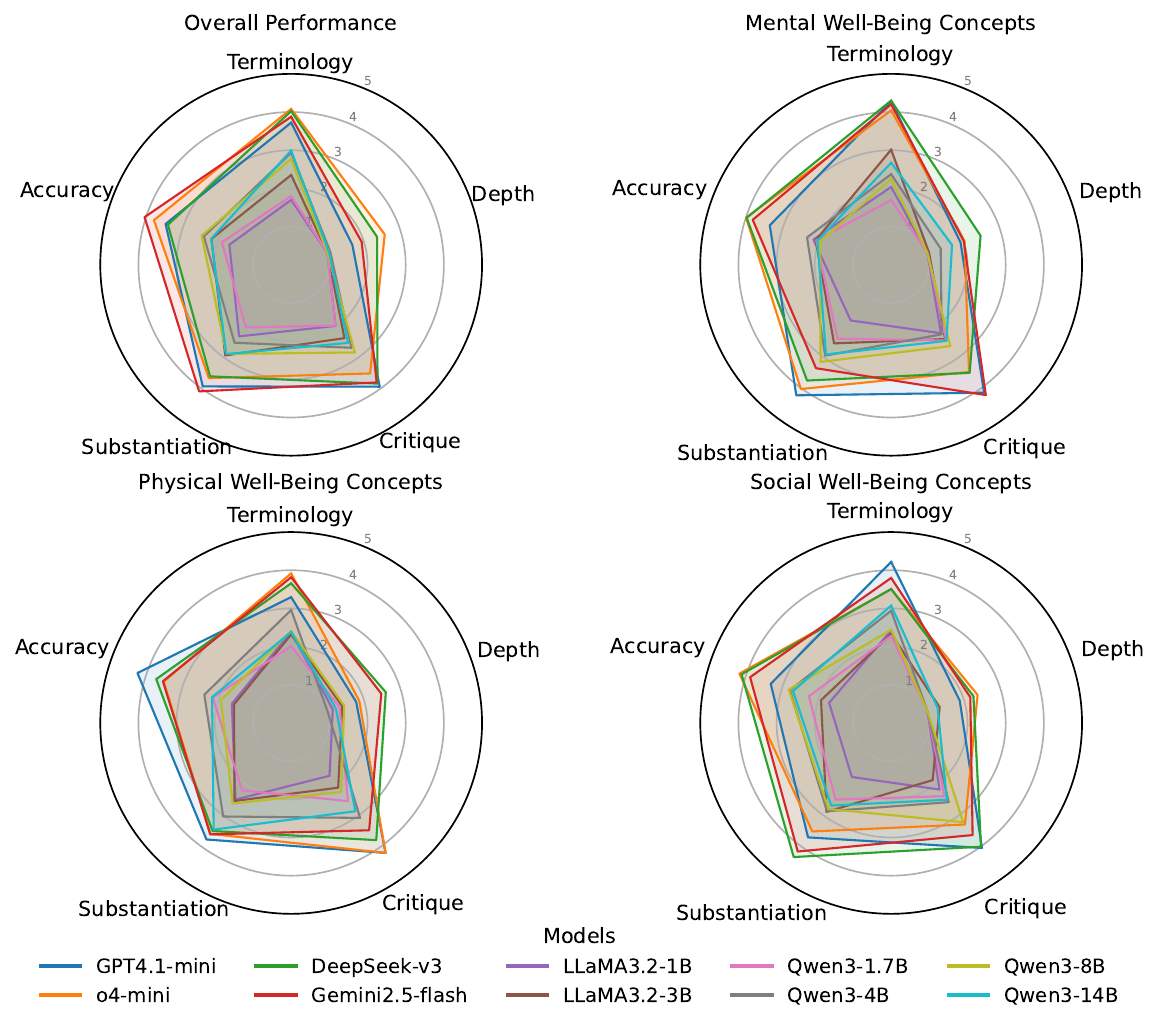}
    \caption{Scores of concept explanations for domain experts.}
    \label{fig:radar_e}
  \end{subfigure}

  \caption{Direct Scoring comparisons of LLMs' well-being concept explanation. In Figure~(\subref{fig:radar_ne}) and (\subref{fig:radar_e}), ``Overall Performance'' is calculated by averaging scores from mental, physical, and social well-being concepts explanations.}
  \label{fig:radar_subfigs}
\end{figure*}

\begin{table*}[ht]
\centering
\begin{tabular}{lcccccccc}
\toprule
\multirow{2}{*}{\textbf{Model}} 
  & \multicolumn{4}{c}{\textbf{General Public}}
  & \multicolumn{4}{c}{\textbf{Domain Expert}} \\
\cmidrule(lr){2-5} \cmidrule(lr){6-9}
  & Mental & Physical & Social & Overall 
  & Mental & Physical & Social & Overall \\
\midrule
\multicolumn{9}{c}{\textit{Larger API-based Models}} \\
\midrule
GPT-4.1-mini        & \underline{88.5} & \textbf{92.3} & 84.7 & \underline{88.5}
                    & 90.3 & 92.7 & 86.9 & 90.0 \\
o4-mini             & 87.4 & 90.8 & \underline{85.3} & 87.8
                    & \textbf{91.8} & \textbf{94.4} & \textbf{88.2} & \textbf{91.5} \\
DeepSeek-v3         & \textbf{89.1} & 91.7 & \textbf{85.9} & \textbf{88.9}
                    & \underline{90.5} & \underline{93.6} & \underline{87.6} & \underline{90.6} \\
Gemini-2.5-flash    & 86.2 & \underline{91.8} & 83.8 & 87.3
                    & 89.2 & 91.5 & 85.3 & 88.7 \\
\midrule
\multicolumn{9}{c}{\textit{Smaller Open-source Models}} \\
\midrule

LLaMA-3.2-1B-Instruct        & 12.4 & 18.7 &  7.5 & 12.9
                    & 14.1 & 22.3 &  9.2 & 15.2 \\
LLaMA-3.2-3B-Instruct        & 35.2 & 55.3 & 25.1 & 38.5
                    & 45.8 & 72.1 & 31.4 & 49.8 \\
Qwen-3-1.7B         & 22.7 & 48.3 & 13.6 & 28.2
                    & 26.5 & 52.4 & 17.2 & 32.0 \\
Qwen-3-8B           & \underline{65.0} & \underline{80.5} & \underline{53.2} & \underline{66.2}
                    & \underline{68.3} & \underline{82.1} & \underline{63.1} & \underline{71.2} \\
Qwen-3-14B          & \textbf{78.4} & \textbf{88.7} & \textbf{65.9} & \textbf{77.7}
                    & \textbf{81.3} & \textbf{90.2} & \textbf{68.4} & \textbf{80.0} \\
Qwen-3-4B (baseline)&  0.0 &  0.0 &  0.0 &  0.0
                    &  0.0 &  0.0 &  0.0 &  0.0 \\
\bottomrule
\end{tabular}
\caption{Comparative Ranking comparisons of LLMs’ well-being concept explanation. All results indicate win rates (\%) against the Qwen-3-4B baseline on the \textit{whole dataset}. \textbf{Bold} and \underline{underline} values indicate the best and second-best results, respectively.}
\label{tab:win_rates}
\end{table*}

To answer \textbf{RQ2}, we conduct comprehensive analyses of the evaluated pretrained LLMs. Our results reveal significant disparities in LLMs' capabilities based on several factors.

\paragraph{Model size effect: larger LLMs are more capable of well-being concept explanation.}
Figure~\ref{fig:radar_subfigs} compares each LLM’s Direct Scoring results across five evaluation principles for the general public (Figure~\ref{fig:radar_ne}) and domain experts (Figure~\ref{fig:radar_e}). In both cases, the four larger API-based LLMs (GPT-4.1-mini, o4-mini, DeepSeek-v3, and Gemini-2.5-flash) form substantially larger radar polygons than the smaller open-source models, indicating a clear scale effect.

In Table~\ref{tab:win_rates}, the top four larger LLMs achieve overall win rates exceeding 87\% for the general public audience and 88\% for the domain expert audience against the baseline model. DeepSeek-v3 emerges as the top performer for the general public with an 88.9\% win rate, while o4-mini leads for the domain expert audience with a 91.5\% win rate. In contrast, the performance of the smaller open-source LLMs scales with parameter count but remains significantly lower. This performance divide is visually confirmed by the radar charts in Figure~\ref{fig:radar_subfigs}, where the Larger LLMs consistently form a large and outer performance polygon, while the smaller models are clustered in a much smaller area, indicating lower scores across all evaluation principles.

\paragraph{Audience effect: generating high-quality well-being concept explanation for domain experts is challenging.}
While LLMs can effectively adapt their concept explanations to the target audience, they are struggling to provide good explanations for \textit{domain experts} with specialized background knowledge. A comparative analysis of explanation for the general public (Figures~\ref{fig:radar_ne}) and domain experts (Figure~\ref{fig:radar_e}) reveals two findings: (1) The quality of concept explanations for domain experts is worse than those generated for the general public, reflecting on the overall smaller radar polygons and \textit{Accuracy} decrease. For example, DeepSeek-v3 falls from 4.72 to 3.41 (–27.8\%), while o4-mini plunges from 4.73 to 3.72 (–21.4\%). This systematic decline indicates that, when asking LLMs to generate explanations for domain experts, they are more likely to hallucinate or generate factually inaccurate details. (2) The performance disparity between smaller and larger LLMs is increasing in expert-oriented concept explanations. This phenomenon can be further confirmed by the higher win rates of domain experts compared to the general public. We speculate that the two findings are possibly due to the limited learning capacity of smaller models when there is a lack of high-quality, professional, and jargon-rich data that would enable more nuanced explanations of well-being concepts.

\paragraph{Well-being category effect: explaining social well-being concepts is more difficult.}
Besides model scale and audience type, 
we observe that \textit{Physical} well‐being explanations (lower‐left quadrants in Figure~\ref{fig:radar_ne} and \ref{fig:radar_e}) achieve the highest overall quality: all four larger LLMs score above 4.5 and nearly 4 in direct scoring on \textit{Accessibility} and \textit{Terminology}, respectively. In contrast, \textit{Social} well‐being explanations show the greatest variability in the radar chart and lowest win rates among the three well-being concept categories (Table~\ref{tab:win_rates}). \textit{Mental} well‐being explanations sit between these extremes: nearly all LLMs show the median win rates among three well-being concept categories (Table~\ref{tab:win_rates}).

\paragraph{Principle-wise analysis: larger LLMs present unified weakness in Utility and Depth despite diverse strengths.}
As shown in Figure~\ref{fig:radar_subfigs}, while the larger LLMs consistently outperform smaller models across all evaluation principles, they exhibit a shared weakness in providing practical advice (\textit{Utility}) for the general public and generating nuanced analyses (\textit{Depth}) for domain experts. At the same time, each of these models demonstrates particular strengths in specific principles. GPT-4.1-mini excels on \textit{Accessibility} and \textit{Terminology}, o4-mini achieves the highest scores for factual \textit{Accuracy} in both settings, DeepSeek-v3 is good at providing clear \textit{Demonstration} and \textit{Concise} explanations, and Gemini-2.5-flash can generate \textit{Accurate} definitions as well as providing evidence and references (\textit{Substantiation}). Although larger LLMs generally perform worse on \textit{Utility} and \textit{Depth}, Gemeni-flash-2.5 and DeepSeek-v3 demonstrate relatively better performances. Based on their overall performance (Figure~\ref{fig:radar_subfigs}), we list the winner for each evaluation principle. 
For the principles of \textit{general public}:
\begin{itemize}[leftmargin=*,itemsep=1pt]
    \item Accuracy: o4-mini and DeepSeek-v3
    \item Accessibility: GPT-4.1-mini
    \item Conciseness: o4-mini and DeepSeek-v3
    \item Demonstration: DeepSeek-v3 and GPT-4.1-mini
    \item Utility: Gemini-2.5-flash and DeepSeek-v3
\end{itemize}
For the principles of \textit{domain experts}:
\begin{itemize}[leftmargin=*,itemsep=1pt]
    \item Accuracy: Gemini-flash-2.5
    \item Terminology: o4-mini and GPT-4.1-mini
    \item Depth: o4-mini
    \item Critique: DeepSeek-v3 and GPT-4.1-mini
    \item Substantiation: Gemini-flash-2.5
\end{itemize}

\subsection{Performances of Fine-Tuned Well-Being Concept Explanation Models (RQ3)}
To respond to \textbf{RQ3}, we fine‐tune the Qwen-3-4B base model using SFT and DPO. We compare their performance back on the same evaluation set. In particular, we begin with a pool of 600 well‐being concepts, split evenly into 300 for training and 300 held out for evaluation. 
For each training concept, we collect four \textit{good} and two \textit{bad} explanations: SFT uses only the \text{good} examples, while DPO uses paired \text{good} and \text{bad} examples. We then apply both SFT and DPO to the Qwen-3-4B model and evaluate all LLMs on the evaluation set using our Direct Scoring and Comparative Ranking.
\begin{table}[t]
\centering
\begin{tabular}{lcc}
\toprule
\textbf{Model} & \textbf{General Public} & \textbf{Domain Expert} \\
\midrule
\multicolumn{3}{c}{\textit{Larger API-based Models}} \\
\midrule
GPT-4.1-mini & 4.17 & 4.00 \\
o4-mini & \underline{4.21} & \underline{4.05} \\
DeepSeek-v3 & \textbf{4.25} & \textbf{4.10} \\
Gemini-2.5-flash & 4.19 & 4.02 \\
\midrule
\multicolumn{3}{c}{\textit{Smaller Open-source Models}} \\
\midrule
LLaMA-1B-Inst. & 1.98 & 1.60 \\
LLaMA-3B-Inst. & 2.72 & 2.38 \\
Qwen-3-1.7B & 2.11 & 1.55 \\
Qwen-3-8B & \underline{2.98} & \underline{2.62} \\
Qwen-3-14B & \textbf{3.26} & \textbf{2.78} \\
\midrule
\multicolumn{3}{c}{\textit{Baseline \& Fine-tuned Models}} \\
\midrule
Qwen-3-4B & 2.74 & 2.47 \\
Qwen-3-4B-SFT & \underline{3.18 (+16.1\%)} & \underline{2.79 (+13.0\%)} \\
Qwen-3-4B-DPO & \textbf{3.25 (+18.6\%)} & \textbf{2.85 (+15.4\%)} \\
\bottomrule
\end{tabular}
\caption{Comparison of Direct Scoring results on the \textit{evaluation set}. Averaged scores of all principles are presented. Relative gains over Qwen-3-4B are shown in parentheses.}
\label{tab:direct_scores_all_with_ft}
\end{table}

\begin{table}[t]
\centering
\begin{tabular}{lcc}
\toprule
\textbf{Model}            & \textbf{General Public} & \textbf{Domain Expert} \\
\midrule
\multicolumn{3}{c}{\textit{Larger API-based Models}} \\
\midrule
GPT-4.1-mini              & \underline{88.1}                     & 89.2                    \\
o4-mini                   & 87.2                     & \textbf{90.7}                    \\
DeepSeek-v3               & \textbf{88.3}                     & \underline{89.8}                    \\
Gemini-2.5-flash          & 87.0                     & 88.1                    \\
\midrule
\multicolumn{3}{c}{\textit{Smaller Open-source Models}} \\
\midrule
LLaMA-1B-Inst.              & 13.5                     & 16.0                    \\
LLaMA-3B-Inst.              & 38.4                     & 52.6                    \\
Qwen-3-1.7B               & 20.7                     & 30.4                    \\
Qwen-3-8B                 & \underline{66.5}                     & \underline{70.5}                    \\
Qwen-3-14B                & \textbf{77.5}                     & \textbf{79.3}                    \\
\midrule
\multicolumn{3}{c}{\textit{Baseline \& Fine-tuned Models}} \\
\midrule
Qwen-3-4B      &  0.0                     &  0.0                    \\
Qwen-3-4B-SFT             & \underline{72.2 (+72.2\%)}       & \underline{81.4 (+81.4\%)}      \\
Qwen-3-4B-DPO             & \textbf{75.9 (+75.9\%)} & \textbf{83.4 (+83.4\%)} \\
\bottomrule
\end{tabular}
\caption{Comparative Ranking results against Qwen-3-4B on the \textit{evaluation set}. Overall win rates (\%) are reported and relative gains over Qwen-3-4B are shown in parentheses.}
\label{tab:comp_rank_eval_with_ft}
\end{table}

\paragraph{Improvements on direct scoring results.}
As shown in Table~\ref{tab:direct_scores_all_with_ft}, both fine‐tuning strategies achieve substantial gains over the pre-trained Qwen-3-4B model. Qwen-3-4B-SFT increases the general public score by 0.44 points (+16.1\%) to 3.18 and the expert score by 0.32 points (+13.0\%) to 2.79, completely outperforming the Qwen-3-4B and 8B and nearly matching the Qwen-3-14B performance. Qwen-3-4B-DPO improves even further, adding 0.51 points (+18.6\%) to 3.25 for the general public and 0.38 points (+15.4\%) to 2.85 for the domain expert.

\paragraph{Improvements on comparative ranking results.}
Table~\ref{tab:comp_rank_eval_with_ft} presents win rate increases on the evaluation set. Qwen-3-4B-SFT achieves the general public win rate of 72.2\% and the expert win rate of 81.4\%, positioning between the larger Qwen-3 variants (8B and 14B). On the other hand, Qwen-3-4B-DPO further increases the general public's win rate to 75.9\% and the expert's to 83.4\%, surpassing Qwen-3-14B for domain expert concept explanations. Although they are still not comparable with larger API-based LLMs, these results demonstrate that both SFT and DPO can bring smaller LLMs' performance up to the level of their larger variants after fine-tuning on our datasets.

\paragraph{DPO generally achieves better performance than SFT.}
Although both fine-tuning approaches significantly improve Qwen-3-4B’s explanation quality, Qwen-3-4B-DPO consistently outperforms Qwen-3-4B-SFT across both Direct Scoring and Comparative Ranking (Table~\ref{tab:direct_scores_all_with_ft} and~\ref{tab:comp_rank_eval_with_ft}). We attribute this to DPO's preference-driven training objective, which directly optimizes the model to prefer higher-quality explanations over lower-quality ones, rather than merely mimicking good examples. Thus, DPO captures more subtle signals from good and bad examples than standard maximum likelihood (i.e., SFT). 

\section{Conclusion and Future Work}
In this paper, we systematically evaluate whether LLMs are ready to explain complex well-being concepts. We build a large-scale dataset of well-being concept explanations, develop a principle-guided evaluation framework, and test the efficacy of the small fine-tuned models using both SFT and DPO. Our findings reveal shared weaknesses of LLMs. We point out that LLM can struggle with factual Accuracy when explaining concepts to experts. Finally, we demonstrate that both SFT and DPO substantially improve smaller models. Future work could explore the efficacy of other tuning techniques, such as Proximal Policy Optimization (PPO)~\cite{schulman2017proximal}, Constrained Policy Optimization (CPO)~\cite{achiam2017constrained}, and Group Relative Policy Optimization (GRPO)~\cite{shao2024deepseekmath}. Moreover, researchers can follow our research pipeline to collect and evaluate more LLM-generated concept explanations for other types of audiences (e.g., K12 students) or from different domains (e.g., physics).

\bibliography{aaai2026}


\end{document}